\documentclass[11pt,a4paper]{article}
\usepackage[hyperref]{naaclhlt2019}
\usepackage{times}
\usepackage{latexsym}
\usepackage{calculator}
\usepackage{subcaption}
\usepackage{xstring}
\usepackage{amsmath}
\usepackage{graphicx}
\usepackage{url}
\usepackage{array}
\usepackage{multirow}
\usepackage{siunitx}
\usepackage{booktabs}
\usepackage{etoolbox}
\newcolumntype{P}[1]{>{\centering\arraybackslash}p{#1}}
\renewrobustcmd{\bfseries}{\fontseries{b}\selectfont}
\renewrobustcmd{\boldmath}{}
\newrobustcmd{\B}{\bfseries}
\usepackage{xcolor}
\usepackage{color}
\usepackage[normalem]{ulem} 

\aclfinalcopy 

\newcommand{\layer}[1]{\IfEqCase{#1}{%
        {1}{1$^{st}$}
        {2}{2$^{nd}$}
        }
}
\newcommand{\unit}[2]{\IfEqCase{#1}{
    {1}{#2\;} 
    {2}{\ADD{#2}{650}{\unitnr}\unitnr\;}
    }
}

\title{The emergence of number and syntax units in LSTM language models}

\author{Yair Lakretz\\Cognitive Neuroimaging Unit\\NeuroSpin center\\91191, Gif-sur-Yvette, France\\{\tt yair.lakretz@gmail.com}\\
\And
	German Kruszewski\\Facebook AI Research\\Paris, France\\ {\tt germank@gmail.com}
	\AND
	Theo Desbordes\\Facebook AI Research\\Paris, France\\{\tt tdesbordes@fb.com}
  \And
	Dieuwke Hupkes\\ILLC, University of Amsterdam\\Amsterdam, Netherlands\\ {\tt d.hupkes@uva.nl}
  \AND
	Stanislas Dehaene\\Cognitive Neuroimaging Unit\\NeuroSpin center\\91191, Gif-sur-Yvette, France\\{\tt stanislas.dehaene@gmail.com}
  \And
	Marco Baroni\\Facebook AI Research\\Paris, France\\{\tt mbaroni@fb.com}}

\date{}

\begin{document}

\maketitle

\begin{abstract}
  Recent work has shown that LSTMs trained on a generic language
  modeling objective capture syntax-sensitive generalizations such as
  long-distance number agreement.  We have however no mechanistic
  understanding of how they accomplish this remarkable feat. Some have
  conjectured it depends on heuristics that do not truly take
  hierarchical structure into account.  We present here a detailed
  study of the inner mechanics of number tracking in LSTMs at the
  single neuron level.  We discover that long-distance number
  information is largely managed by two ``number units''.
  Importantly, the behaviour of these units is partially controlled by
  other units independently shown to track syntactic structure.  We
  conclude that LSTMs are, to some extent, implementing genuinely
  syntactic processing mechanisms, paving the way to a more general
  understanding of grammatical encoding in LSTMs.
\end{abstract}

\section{Introduction}

In the last years, recurrent neural networks (RNNs), and particularly
long-short-term-memory (LSTM) architectures
\cite{Hochreiter:Schmidhuber:1997}, have been successfully applied to
a variety of NLP tasks. This has spurred interest in whether these
generic sequence-processing devices are discovering genuine structural
properties of language in their training data, or whether their
success can be explained by opportunistic surface-pattern-based
heuristics.

Until now, this debate has mostly relied on ``behavioural'' evidence:
The LSTM had been treated as a black box, and its capacities had been indirectly
inferred by its performance on linguistic tasks. In this study, we took a
complementary approach inspired by neuroscience: We thoroughly investigated
the inner dynamics of an LSTM language model performing a number agreement task,
striving to achieve a mechanistic understanding of how it accomplishes
it. We found that the LSTM had specialized two ``grandmother'' cells
\cite{Bowers:2009} to carry number features from the subject to the
verb across the intervening material.\footnote{In the neuroscientific
  literature, ``grandmother'' cells are (sets of) neurons coding for
  specific information, e.g., about your grandmother, in a
  non-distributed manner.} Interestingly, the LSTM also
possesses a more distributed mechanism to predict number when subject
and verb are close, with the grandmother number cells only playing a
crucial role in more difficult long-distance cases. Crucially, we
independently identified a set of cells tracking syntactic structure,
and found that one of them encodes the presence of an embedded phrase
separating the main subject-verb dependency, and has strong efferent
connections to the long-distance number cells, suggesting that the
network relies on genuine syntactic information to regulate
agreement-feature percolation.

Our analysis thus provides direct evidence for the claim that LSTMs
trained on unannotated corpus data, despite
lacking significant linguistic priors, learn to perform
structure-dependent linguistic operations. In turn, this suggests that
raw linguistic input and generic memory mechanisms, such as those
implemented in LSTMs, may suffice to trigger the induction of
non-trivial grammatical rules.

\section{Related work}

Starting with the seminal work of \newcite{Linzen:etal:2016}, a long-distance number agreement task has emerged as a standard way to probe the syntactic capabilities of neural language models.
In the number agreement task, a model is asked to predict the verb in a sentence where the subject and main verb are separated by one or more intervening nouns (``the \textbf{boy} near the \textit{cars} \textbf{greets}\ldots'') and evaluated based on how often it predicts the right verb form.

Following mixed initial results by Linzen and colleagues and \newcite{Bernardy:Lappin:2017}, \newcite{Gulordava:etal:2018} and
\newcite{Kuncoro:etal:2018a} have robustly established that LSTM language models
achieve near-human performance on the agreement task. While Gulordava and
colleagues provided some evidence that the LSTMs are relying on
genuine syntactic generalizations, \newcite{Kuncoro:etal:2018b} and
\newcite{Linzen:Leonard:2018} suggested that the LSTM achievements
can, at least in part, be accounted by superficial heuristics (e.g., ``percolate the number of the first noun in a sentence''). Other
recent work has extended syntax probing to other phenomena such as
negative polarity items and island constraints
\cite{Chowdhury:Zamparelli:2018,jumelet2018language,marvin2018targeted,wilcox2018rnn}.

While \newcite{Linzen:etal:2016} presented intriguing qualitative data
showing cells that track grammatical number in a network directly trained on
the agreement task, most of the following work focused on testing the
network output behaviour, rather than on understanding how the latter
follows from the inner representations of the network. Another research line
studied linguistic processing in neural networks through `diagnostic classifiers', that is, classifiers trained to
predict a certain property from network activations
\cite[e.g.,][]{gelderloos2016phonemes,Adi:etal:2017,alain2017understanding,Hupkes:etal:2017}. This approach may give insight into which information is encoded by the
network in different layers or at different time points, but it only
provides indirect evidence about the specific mechanics of linguistic
processing in the network.

Other studies are closer to our approach in that they attempt to
attribute function to specific network cells, often by means of
visualization
\cite{Karpathy:etal:2016,li2016visualizing,tang2017memory}. \newcite{Radford:etal:2017},
for example, detected a ``sentiment'' grandmother cell in a
language-model-trained network.  \newcite{Kementchedjhieva:Lopez:2018}
recently found a character-level RNN to track morpheme boundaries in a
single cell. We are however not aware of others studies systematically
characterizing the processing of a linguistic phenomenon at the level of
RNN cell dynamics, as is the attempt in the study hereby presented.

\section{Setup}\label{sec:the_data}

\paragraph{Language model}\label{ssec:lstm_lm}
We study the pretrained LSTM language model made available by
\newcite{Gulordava:etal:2018}. This model is composed of a
650-dimensional embedding layer, two 650-dimensional hidden layers,
and an output layer with vocabulary size 50,000.
The model was trained on Wikipedia data, without fine-tuning for number
agreement, and obtained perplexity close to state of the art in the experiments of Gulordava et al.\footnote{Key findings reported below were also replicated with the same model trained with different initialization seeds and variations with different hyper-parameters.}

\paragraph{Number-agreement tasks}
\begin{table}[tb]
  \centering
  \begin{footnotesize}
  \begin{tabular}{l@{\hskip1pt}l}
    \B Simple & the \textbf{boy} \textbf{greets} the guy\\
    \B Adv & the \textbf{boy} probably \textbf{greets} the guy\\
    \B 2Adv & the \textbf{boy} most probably \textbf{greets} the guy\\
    \B CoAdv &  the \textbf{boy} openly and deliberately \textbf{greets} the guy\\
    \B NamePP & the \textbf{boy} near Pat \textbf{greets} the guy\\
    \B NounPP & the \textbf{boy} near the car \textbf{greets} the guy\\
    \B NounPPAdv & the \textbf{boy} near the car kindly \textbf{greets} the guy\\
  \end{tabular}
  \end{footnotesize}
  \caption{NA tasks illustrated by representative
    singular sentences.}
  \label{tab:data-sets}
\end{table}

We complement analysis of the naturalistic, corpus-derived
number-agreement test set of \newcite{Linzen:etal:2016}, in the
version made available by \newcite{Gulordava:etal:2018}, with
synthetically generated data-sets. Each synthetic number-agreement
task (NA-task) instantiates a fixed syntactic structure with varied
lexical material, in order to probe subject-verb number agreement in
controlled and increasingly challenging setups.\footnote{We exclude,
  for the time being, agreement across a relative clause, as it comes
  with the further complication of accounting for the extra agreement
  process taking place inside the relative clause.} The different
structures are illustrated in Table \ref{tab:data-sets}, where all forms are in the singular. Distinct sentences were randomly
generated by selecting words from pools of 20
subject/object nouns, 15 verbs, 10 adverbs, 5 prepositions, 10 proper
nouns and 10 location nouns. The items were selected so that their
combination would not lead to semantic anomalies. For each NA-task, we
generated singular and plural versions of each sentence. We refer to
each such version as a \textit{condition}. For NA-tasks that have
other nouns occurring between subject and main verb, we also
systematically vary their number, resulting in two \textit{congruent}
and two \textit{incongruent} conditions. For example, the NounPP
sentence in the table illustrates the congruent SS (singular-singular)
condition and the corresponding sentence in the incongruent PS
(plural-singular) condition is: ``the \emph{boys} near the \emph{car}
\emph{greet} the guy''. For all NA-tasks, each condition consisted of
600 sentences

\paragraph{Syntactic depth data-set} We probed the implicit
syntax-parsing abilities of the model by testing whether its representations
predict the syntactic depth of the words they process. Following
\newcite{Nelson:etal:2017}, this was operationalized as predicting the
number of open syntactic nodes at each word, given the canonical
syntactic parse of a sentence.  We generated a data-set of sentences with
unambiguous but varied syntactic structures and annotated them with the number of open nodes at each word. For example:
``Ten$_1$ really$_2$ ecstatic$_3$ cousins$_3$ of$_4$ four$_5$ teachers$_6$
are$_2$ quickly$_3$ laughing$_4$", where indexes show the corresponding
number of open nodes.
Since syntactic depth is
naturally correlated with the position of a word in a sentence, we
used a data-point sampling strategy to de-correlate these factors. For
each length between 2 and 25 words, we randomly generated 300
sentences. From this set, we randomly picked examples uniformly
covering all possible position-depth combinations within the 7-12
position and 3-8 depth ranges. The final data-set contains 4,033
positions from 1,303 sentences.\footnote{All our data-sets are available at: \url{https://github.com/FAIRNS/Number_and_syntax_units_in_LSTM_LMs}.}

\section{Experiments}\label{sec:results}
To successfully perform the NA-task, the LSTM should: (1) encode and store the grammatical number of the subject; and (2) track the main subject-verb syntactic dependency. The latter information is important for identifying the time period during which subject number should be stored, output and then updated by the network. This section describes the `neural circuit' that encodes and processes this information in the LSTM.

\begin{center}
\begin{table}[t]
\centering
\begin{tabular}{|P{2.3cm}|P{0.4cm}||P{0.6cm}|P{0.6cm}|P{0.6cm}|}
\hline
    \B \multirow{2}{*}{NA task} & \B \multirow{2}{*}{C} & \multicolumn{2}{c|}{\B Ablated}& \multirow{2}{*}{\B Full} \\
    \cline{3-4}
    & & \B \textbf{\unit{2}{126}} & \B \textbf{\unit{2}{338}} & \\
\hline

Simple & S & - &  - &  \textcolor{red}{100} \\

Adv & S & - &  - &  \textcolor{red}{100} \\

2Adv & S & - &  - &  \textcolor{red}{99.9} \\

CoAdv & S & - &  \textcolor{red}{82} &  \textcolor{red}{98.7} \\

namePP & SS & - &  - &  \textcolor{red}{99.3} \\

nounPP & SS & - &  - &  \textcolor{red}{99.2} \\

nounPP & SP &  - &  \textcolor{red}{54.2} &  \textcolor{red}{87.2} \\

nounPPAdv & SS &  - &  - & \textcolor{red}{99.5} \\

nounPPAdv & SP &  - &  \textcolor{red}{54.0} & \textcolor{red}{91.2} \\

\hline
Simple & P &  - &  - &  \textcolor{blue}{100} \\

Adv & P &  - &  - &  \textcolor{blue}{99.6} \\

2Adv & P & - &  - &   \textcolor{blue}{99.3} \\

CoAdv & P &  \textcolor{blue}{79.2} &  - &   \textcolor{blue}{99.3} \\

namePP & PS & \textcolor{blue}{39.9} &  - &   \textcolor{blue}{68.9} \\

nounPP & PS &  \textcolor{blue}{48.0} & - &   \textcolor{blue}{92.0} \\

nounPP & PP &  \textcolor{blue}{78.3} & - &   \textcolor{blue}{99.0} \\

nounPPAdv & PS & \textcolor{blue}{63.7} &  - &   \textcolor{blue}{99.2} \\

nounPPAdv & PP & - &  - &   \textcolor{blue}{99.8} \\

\hline

\B Linzen & \B - &   75.3 &  - &  93.9 \\
\hline

\end{tabular}
\caption{Ablation-experiments results: Percentage accuracy in all NA-tasks. Full: non-ablated model, C: condition, S: singular, P: plural. Red: Singular subject, Blue: Plural subject. Performance reduction less than 10\% is denoted by `-'.  \label{tab:ablation-results}}
\end{table}
\end{center}

\subsection{Long-range number units}\label{subsec:ablation}
We first tested the performance of the LSTM on the Linzen's data and on the NA-tasks in Table
\ref{tab:data-sets}. Following
\newcite{Linzen:etal:2016} and later work, we computed the likelihood
that the LSTM assigns to the main verb of each sentence given the
preceding context and compared it to the likelihood it assigns to the
wrong verb inflection. Accuracy in a given condition was measured as the proportion of sentences in this condition for which the model assigned a higher likelihood to the correct verb form than to the wrong one.

Network performance is reported in Table \ref{tab:ablation-results}
(right column -- `Full'). We first note that our results on the Linzen NA-task confirm those reported in \newcite{Gulordava:etal:2018}. For the other NA-tasks, results show that some tasks and
conditions are more difficult than others. For example, performance on
the Simple (0-distance) NA-task is better than that on the Co-Adv
NA-task, which in turn is better than that of the nounPP
tasks. Second, as expected, incongruent conditions (the
number-mismatch conditions of namePP, nounPP and nounPPAdv) reduce
network performance. Third, for long-range dependencies, reliably
encoding singular subject across an interfering noun is more difficult
than a plural subject: for both nounPP and nounPPAdv, PS is easier
than SP. A possible explanation for this finding is that in English the plural form is
almost always more frequent than the singular one, as the latter only
marks third person singular, whereas the former is identical to the
infinitive and other forms. Thus, if the network reverts to unigram
probabilities, it will tend to prefer the plural. 

\paragraph{Looking for number units through ablation} Number
information may be stored in the network in either a local, sparse, or
a distributed way, depending on the fraction of active units that
carry it.  We hypothesized that if the network uses a local or sparse
coding, meaning that there's a small set of units that encode number
information, then ablating these units would lead to a drastic
decrease in performance in the NA-tasks.  To test this, we ablated
each unit of the network, one at a time, by fixing its activation to zero,
and tested on the NA-tasks.

Two units were found to have exceptional effect on network performance
(Table \ref{tab:ablation-results}, \unit{2}{126} and \unit{2}{338}
columns).\footnote{Units 1-650 belong to the first layer, 651-1300 to
  the second. All units detected by our analyses come from the latter.} Ablating them reduced network performance by more than 10\%
across various conditions, and, importantly, they were the only units
whose ablation consistently brought network performance to around
chance level in the more difficult incongruent conditions of the
namePP, nounPP and nounPPAdv
tasks. 

Moreover, the ablation effect depended on the grammatical number of the subject: ablating \unit{2}{126} significantly reduced
network performance only if the subject was plural (P, PS or PP conditions) and \unit{2}{338}
only if the subject was singular (S, SP or SS conditions). In what follows, we will therefore
refer to these units as the `plural' and `singular' units, respectively,
or long-range (LR) number units when referring to both. Finally, we note that although the Linzen NA-task contained mixed stimuli from many types of conditions, the plural unit was found to have a substantial effect on average on network performance. The singular unit didn't show a similar effect in this case, which highlights the importance of using carefully crafted stimuli, as in the nounPP and nounPPAdv tasks, for understanding network dynamics. Taken
together, these results suggest a highly local coding scheme of
grammatical number when processing long-range dependencies.

\paragraph{Visualizing gate and cell-state dynamics}\label{subsec:gate-dynamics}
To understand the functioning of the number units, we now look
into their gate and state dynamics during sentence processing. We
focus on the nounPP NA-task, which is the simplest NA-task that includes a
long-range dependency with an interfering noun, in both SP and PS
conditions.

Recall the standard LSTM memory update and output rules \cite{Hochreiter:Schmidhuber:1997}:
\begin{align} 
    C_t &= f_t\circ C_{t-1} + i_t\circ \widetilde{C}_t \label{eq:update-rule} \\
     h_t &= o_t\circ \tanh(C_t) \label{eq:output},
\end{align}
where $f_t, i_t, o_t \in (0,1)$ are gating scalars computed by the network, and $\widetilde{C}_t \in (-1, 1)$ is an update candidate for cell value.

\begin{figure*}[ht]
    \centering
    \begin{subfigure}{\textwidth}
            \centering
            \includegraphics[width=0.4\linewidth]{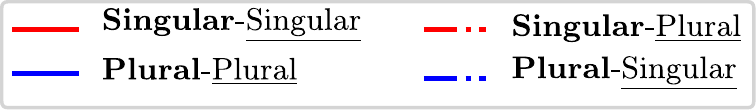}
    \end{subfigure}
    \bigskip
    \begin{subfigure}{0.45\textwidth}
            \centering
            \includegraphics[width=\linewidth]{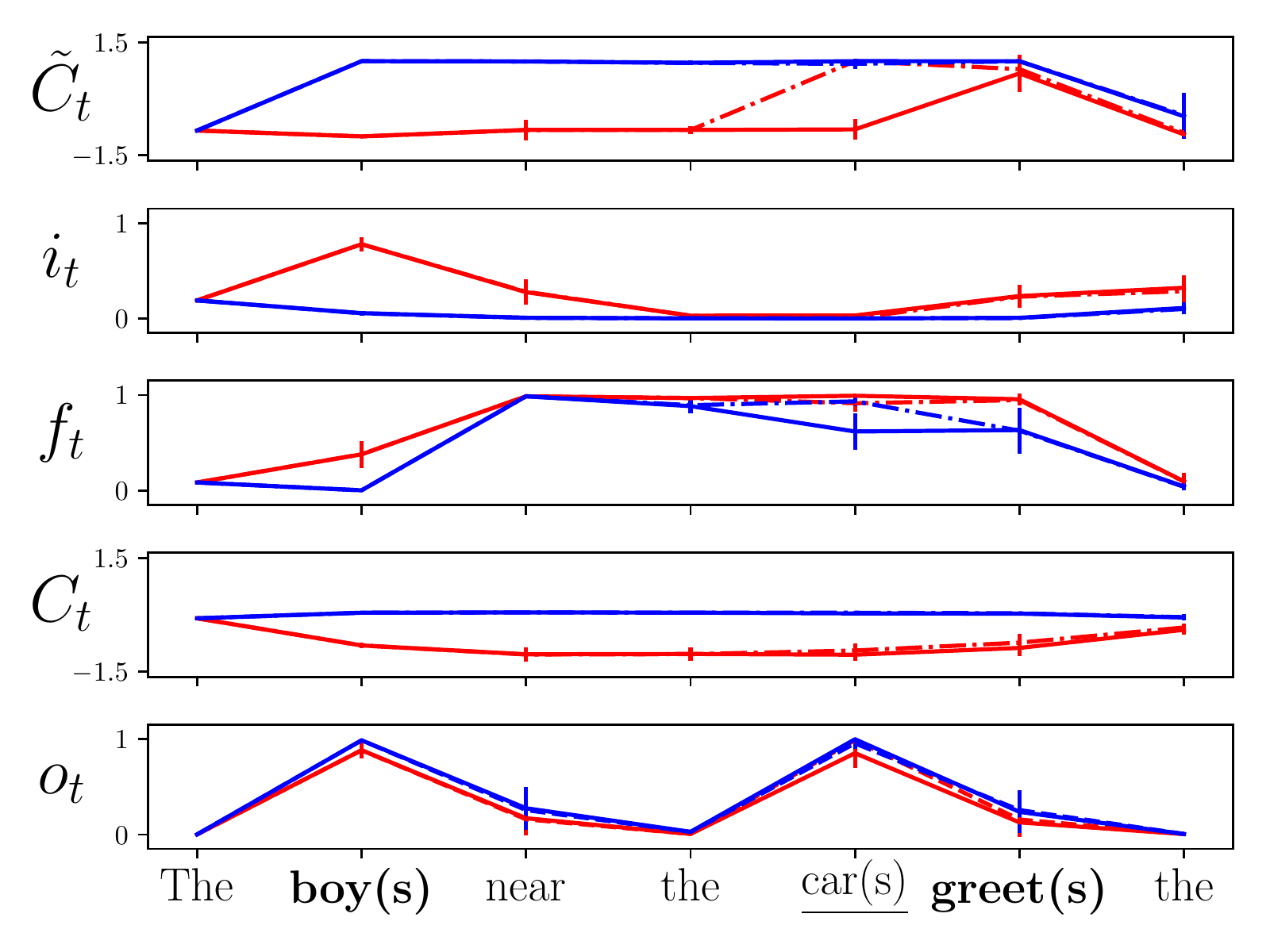}
            \subcaption{\unit{2}{338} (singular)}
    \label{fig:singular-unit}
    \end{subfigure}
    \begin{subfigure}{0.45\textwidth}
            \centering
            \includegraphics[width=\linewidth]{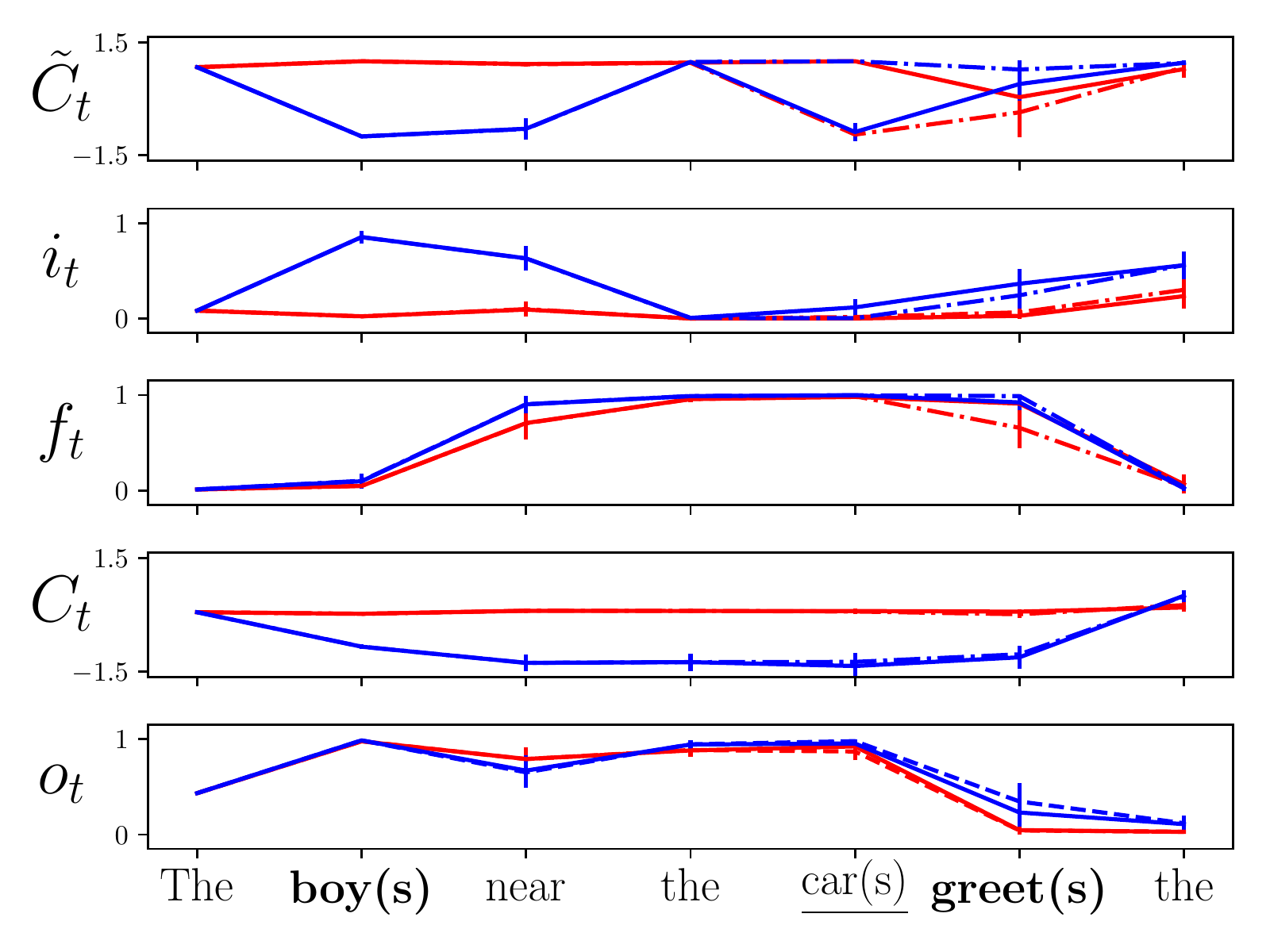}
            \subcaption{\unit{2}{126} (plural)}
    \label{fig:plural-unit}
    \end{subfigure}
    \begin{subfigure}{0.45\textwidth}
            \centering
            \includegraphics[width=\linewidth]{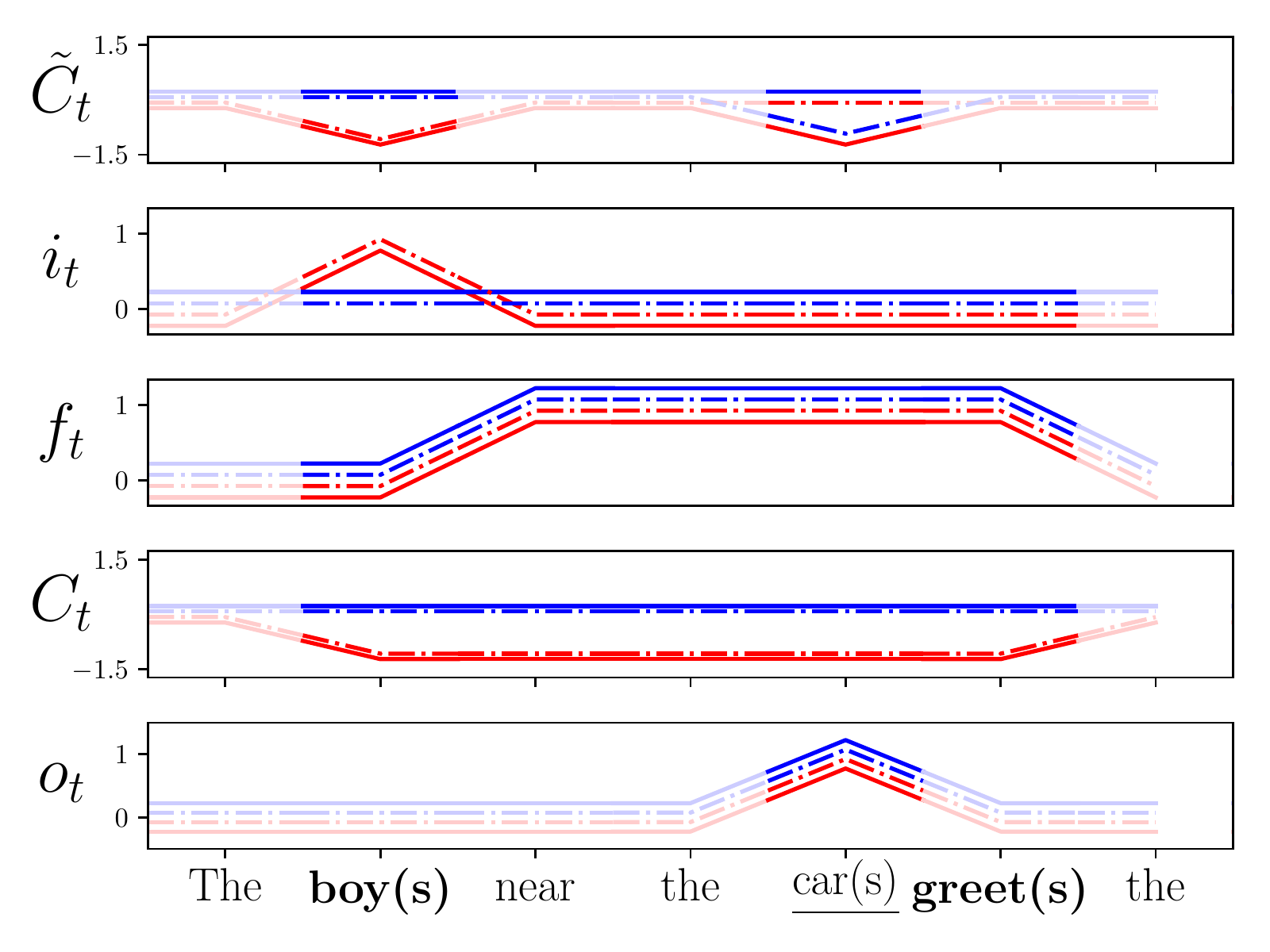}
            \subcaption{Prediction (singular)}
    \label{fig:cartoon}
    \end{subfigure}
    \begin{subfigure}{0.45\textwidth}
        \centering
        \includegraphics[height=4.3cm]{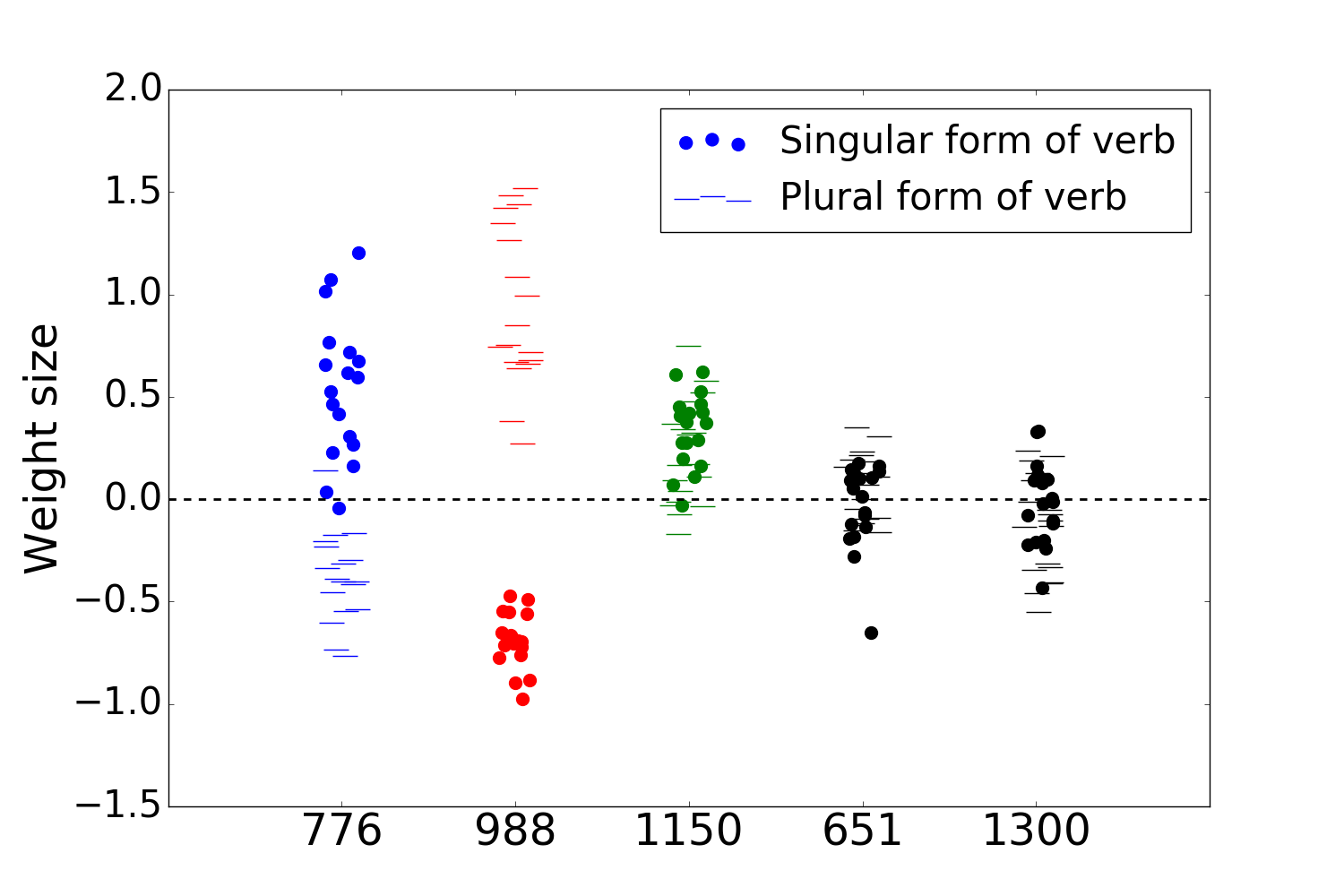}
        \caption{Efferent weights of the LR-units (\unit{2}{126} and \unit{2}{338}), the syntax unit (\unit{2}{500}; section \ref{sec:syntax-units}) and two arbitrary units (\unit{2}{1} and \unit{2}{650}).}
        \label{fig:output-weights}
    \end{subfigure}
    \caption{(a) to (c) -- Cell and gate activations during processing of sentences with a prepositional phrase between subject and verb. Values in (\subref{fig:singular-unit}) and (\subref{fig:plural-unit}) are averaged across all condition sentences, with error bars showing standard deviations. (\subref{fig:output-weights}) -- Efferent weights of specific units at the output layer to singular and plural verb forms.}
\end{figure*}

Consider now how a number unit may reliably encode and store subject
number across interfering nouns.  Figure~\ref{fig:cartoon} exemplifies
this for a singular unit, showing the desired gate and cell
dynamics. The four conditions are represented with separated curves -
red for singular subject, blue for plural, and dashed lines for
incongruent conditions. Gate and cell activity at time points
unrelated to solving the NA-task are masked with white, as we do not
make precise predictions for them.

The update rule of the LSTM cell has two terms
(Eq.~\ref{eq:update-rule}).\footnote{We abuse notation here, using the
  symbols denoting whole layers in equations (\ref{eq:update-rule}) and
  (\ref{eq:output}) to denote the components of single cells.} In the
first, $f_t \circ{} C_{t-1}$, the forget gate controls whether to keep
the previous cell content ($f_t=1$: perfect remembering) or forget it
($f_t=0$: complete forgetting). In the second,
$i_t\circ{} \tilde{C}_t$, the input gate controls whether the
information currently presented to the network, as encoded by
$\tilde{C}_t$, should be written onto the cell ($i_t=1$: full access)
or not ($i_t=0$). The singular unit can thus use these gates to
reliably store number information across long-range
dependencies. Specifically, the unit can (enumeration follows the same order as the panels in Figure~\ref{fig:cartoon}): (1) encode subject
number via $\tilde{C}_{t_{subject}}$ with different values for
singular and plural; (2) open the input gate \textit{only} when a
singular subject is presented ($i_{t_{subject}} = 1$ in red curves \textit{only}) and protect it from interfering nouns ($i_t=0, t_{subject}<t<t_{verb}$); (3) at the same time,
clear the cell from previously stored information
($f_{t_{subject}}=0$) and then store subject number across the entire
dependency ($f_t=1, t_{subject}<t<t_{verb}$); (4) this will result in
stable encoding of subject number in the cell $C_t$ throughout the
dependency; (5) finally, output subject number at the right moment,
when predicting the verb form ($o_{t_{verb}-1}=1$)
(Eq.~\ref{eq:output}).

Figures \ref{fig:singular-unit} and \ref{fig:plural-unit} present the actual gate and cell dynamics of the singular and plural units. Both units follow the general solution for reliable number storage described above. Note that for $\tilde{C}_t$ and $i_t$, and as a result also for $C_t$, the plural unit `mirrors' the singular unit with respect to subject number (red curves of PP and PS vs. blue curves of SS and SP). This is in accordance with the results of the ablation experiments, which showed that ablating these units had an effect that depended on the grammatical number of the subject (Table \ref{tab:ablation-results}). This provides complementary support for the identification of these units as `singular' and `plural'.

A single divergence between the solution depicted in Figure \ref{fig:cartoon} and the actual dynamics of the number units is that input gate activity is smaller, but not zero, at the time step immediately following the subject. One speculative explanation is that this might be useful to process compound nouns. In these cases, subject number information is stored with the second noun, whereas in the case of simple nouns there is no `risk' of encountering an interfering noun immediately after the subject, making the delay in closing the gate safe.


The singular and plural units had emerged at the second layer of the network. This seems appropriate since number information needs to be directly projected to the output layer for correct verb-form prediction. Moreover, number-unit output should be projected differently to singular and plural verb forms in the output layer, only increasing activity in output units representing the suitable form. For example, for the singular unit, since singular subjects are encoded with a negative value ($C_{t_{verb}-1}<-1$ in figure \ref{fig:singular-unit}), the more negative its efferent weights to singular verb forms in the output layer, the higher the probabilities of these verb forms would be. Figure \ref{fig:output-weights} shows the efferent weights of the LR-number units to all verbs in our data-sets. We found that, indeed, the efferent weights to the singular and plural verb forms are segregated from each other, with weight signs that correspond to the negative encoding of subject number used by both singular and plural units. Two other arbitrary units, \unit{2}{1} and \unit{2}{650}, and the syntax unit \unit{2}{500} to be described below (Section \ref{sec:syntax-units}) do not have segregated efferent weights to verb forms, as expected.

\subsection{Short-range number information}
\begin{figure}
    \centering
    \includegraphics[height=4cm, width=8cm]{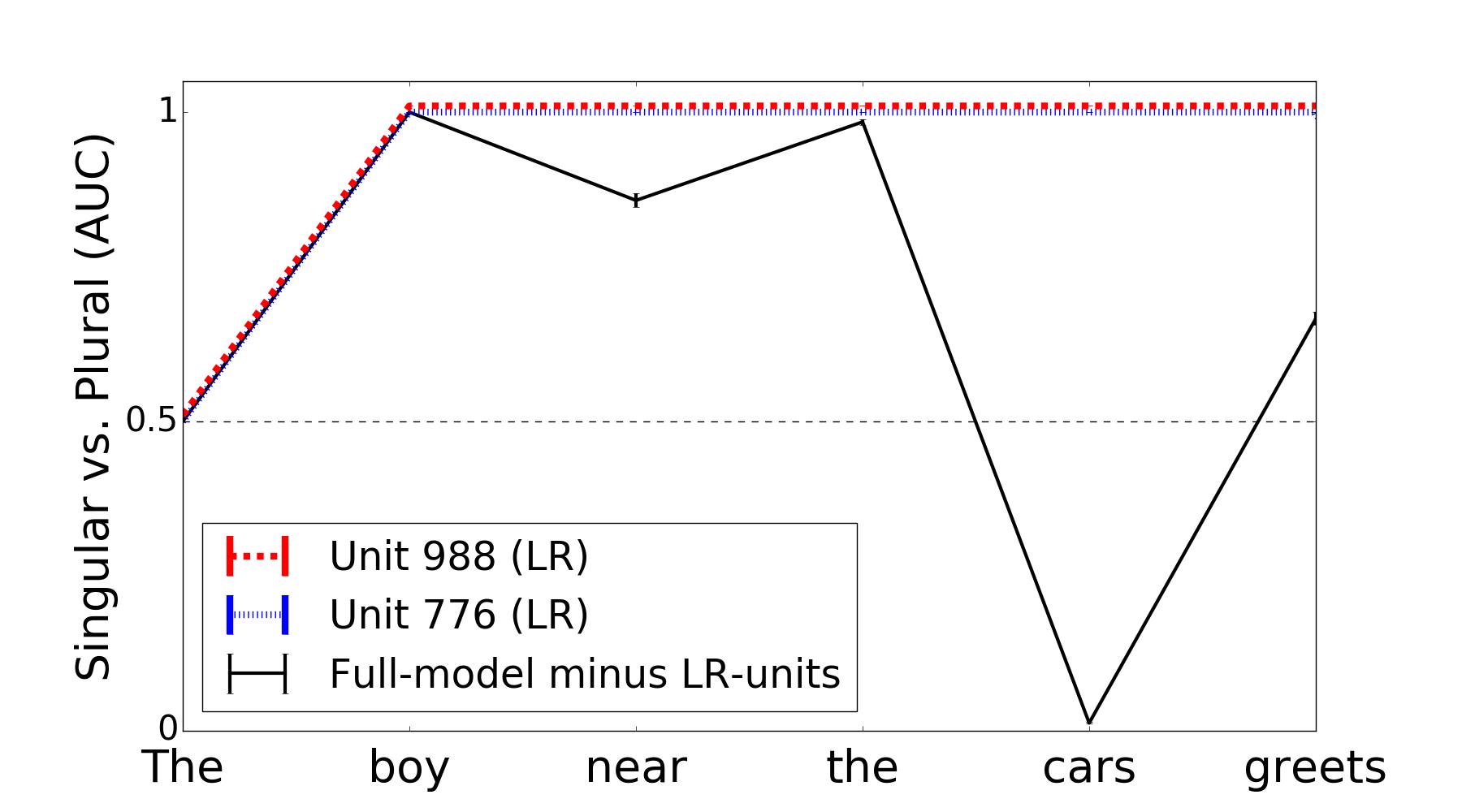}
    \caption{Generalization across time of subject-number prediction. Error bars represent standard deviations across cross-validation splits.}
    \label{fig:GAT}
\end{figure}

Performance on the easier NA-tasks (Simple, Adv, 2Adv) was not
impaired by single-unit ablations. This suggests that number may be
encoded also elsewhere in the network, perhaps via a more distributed
code. To verify this, we tested whether subject number can be decoded from the whole
pattern of activities in the network (excluding the two LR-number units)
and whether this decoding is stable across time \cite[see][for similar
observations and related methods]{Giulianelli:etal:2018}. We expected
this distributed activity to track number in a
small time window after the subject, but, unlike the LR-number units,
to be affected by incongruent intervening nouns.

We trained a linear model to predict the grammatical number of the
subject from network activity in response to the presentation of the
subject, and tested its prediction on test sets from all time points
\cite{King:Dehaene:2014}, in incongruent conditions only of the nounPP
task. We used Area under of Curve (AUC) to evaluate model
performance. Figure \ref{fig:GAT} shows decoding across time of
subject number from cell activity of each number unit separately and
from cell activity of the entire network without these two units
(`Full model minus LR-units'). Results show that number information
can be efficiently decoded from other units in the network, and that
this information can be carried for several time steps (relatively
high AUC up to the second determiner). However, the way in which these
units encode number is sensitive to the last encountered noun, with
AUC decreasing to zero around the second noun (`cars'), whereas test
performance of the models trained on cell activity of the LR-number
units is consistently high. This confirms that number prediction is
supported both by the LR-number units, and by distributed activation
patterns of other short-range (SR) number units. The latter, however,
are not syntax-sensitive, and simply encode the number of the last
noun encountered.

\begin{figure*}[ht!]
    \begin{subfigure}{0.32\textwidth}
            \centering
            \includegraphics[width=\linewidth, height=1.9cm]{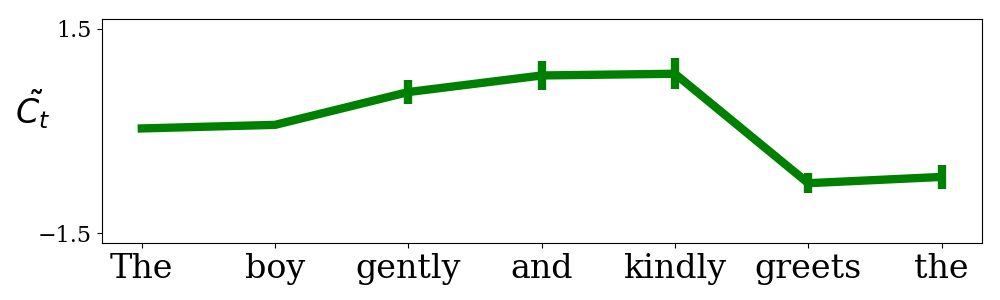}
            \caption{2Adv}
            \label{fig:syntax-unit-2Adv}
    \end{subfigure}
    \begin{subfigure}{0.32\textwidth}
            \centering
            \includegraphics[width=\linewidth, height=1.9cm]{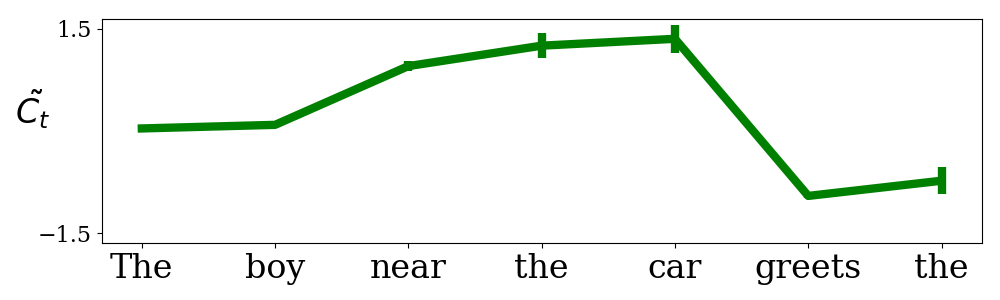}
            \caption{nounPP}
            \label{fig:syntax-unit-nounpp}
    \end{subfigure}
    \begin{subfigure}{0.32\textwidth}
            \centering
            \includegraphics[width=\linewidth, height=1.9cm]{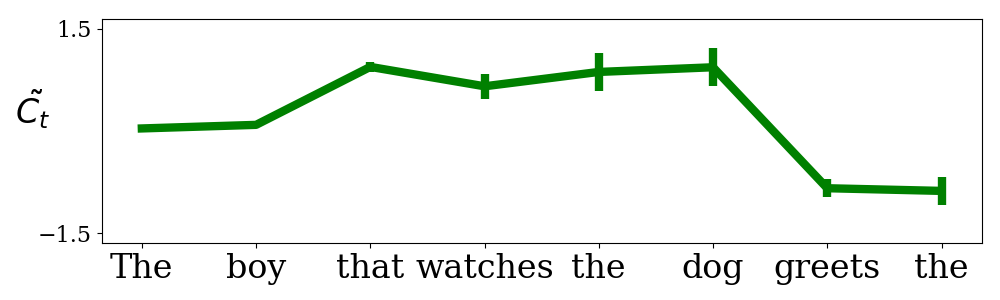}
            \caption{subject relative}
            \label{fig:syntax-unit-subjrel}
    \end{subfigure}
    \begin{subfigure}{\textwidth}
            \centering
            \includegraphics[width=\linewidth, height=2cm]{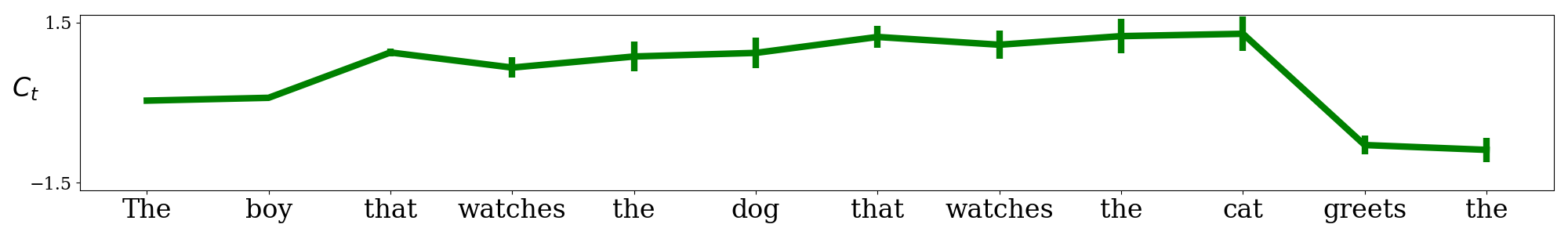}
            \caption{Two embeddings with subject relatives}
            \label{fig:syntax-unit-double-subjrel}
    \end{subfigure}
\caption{Cell activity of syntax unit \unit{2}{500} while processing various syntactic structures. Values averaged across all stimuli in an NA-task, with error bars representing standard deviations. Relative clause NA-task stimuli were specifically generated for this visualization.}
\end{figure*}

A full description of the SR-number units is beyond our scope. However, we note that 10 SR-number units in the second layer of the
network were identified, which had efferent weights with a similar segregated
structure as that of the LR units (Figure
\ref{fig:output-weights}). These units were indeed sensitive to the last
encountered noun: subject number could be decoded from single-unit cell activity
during its presentation (AUC$>0.9$), but activity `swaps' once an
interfering noun appears (i.e., AUC decreases to zero in a
generalization-across-time analysis). Finally, to validate the role of SR-number units in encoding number for easier NA-tasks, we ablated both SR and LR number
units (12 in total) or SR units only (10 in total) and evaluated network performance on these NA-tasks. Both experiments resulted in a significant reduction in task performance compared to 1,000 random equi-size ablations ($p<0.01$ in all `easier' tasks).

Intriguingly, we observed qualitatively that LR units are almost
always making the right prediction, even when the network predicts the
wrong number. The wrong outcome, in such cases, might be due to
interference from the syntax-insensitive SR units. We leave the study of LR-SR
unit interplay to future work.

\subsection{Syntax units}
\label{sec:syntax-units}

\begin{figure*}[ht]
    \centering
    \begin{subfigure}{0.49\textwidth}
            \centering
            \includegraphics[height=3.3cm,width=\textwidth]{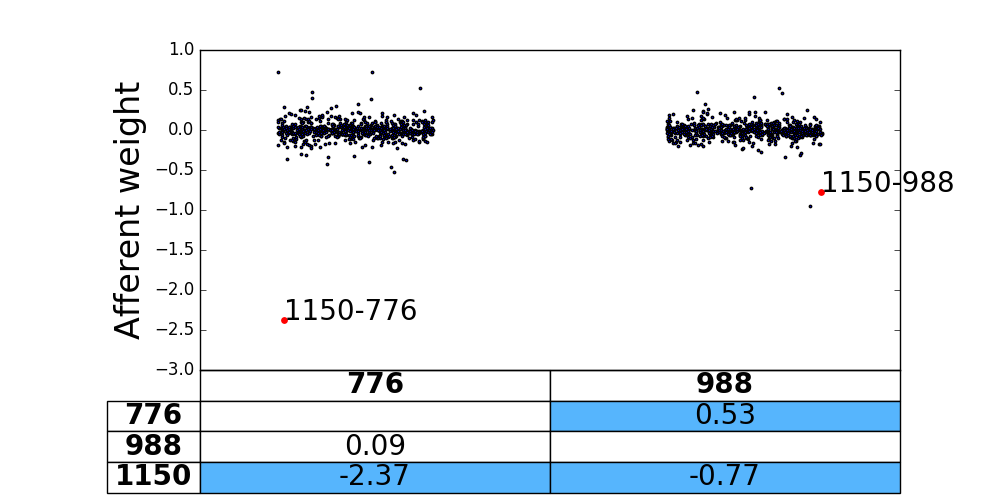}
            \caption{Input gate}
            \label{fig:interaction-input}
    \end{subfigure}
    \begin{subfigure}{0.49\textwidth}
           \centering
          \includegraphics[height=3.3cm,width=\textwidth]{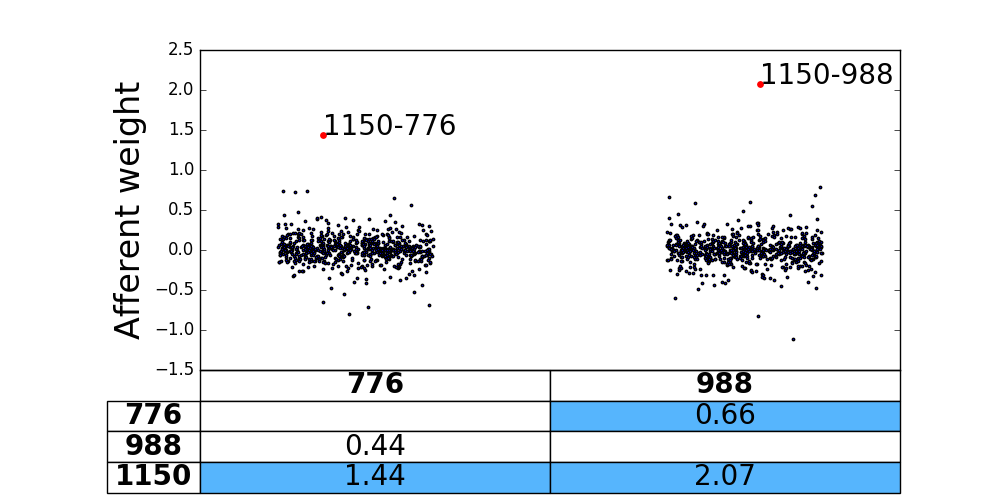}
          \caption{Forget gate}
          \label{fig:interaction-forget}
    \end{subfigure}
    \caption{Connectivity among the syntax unit \unit{2}{500} and
      LR-number units \unit{2}{126} and \unit{2}{338}. Projecting
      units are on the table rows. Blue background highlights outlier
      values ($|z-score|>3$). Weights from the syntax unit are marked in red
      and are explicitly labeled in the plots, which show the overall distributions of afferent weights to each number unit.}
\label{fig:interaction}
\end{figure*}

We saw how the input and forget gates of the LR-number units control the flow of subject-number information. It remains unclear, however, how the dynamics of these gates are controlled by the network. We hypothesized that other units in the network may encode information about the syntactic structure of the sentence, and thus about the subject-verb dependency. These units could then control and coordinate the opening and closing of the input and forget gates of the number units. 

To identify such 'syntax' units, we tested from which units syntactic information can be efficiently decoded. We used depth of the syntactic tree as a proxy for syntactic structure \cite{Nelson:etal:2017} and trained an L2-regularized regression model to predict syntactic tree-depth from the hidden-state activity of all units. In all experiments, we used the data presented in Section \ref{sec:the_data} above and performed a nested 5-fold cross-validation procedure. Word frequency, which was added as a covariate to the model, had a negligible effect on the results. Syntactic tree-depth was found to be efficiently decodable from network activity ($R^2_{test-set}=0.85\pm0.009$; covariate-corrected). A small subset of `syntax' units had relatively high weights in the regression model (mean weight = $7.6\times{}10^{-4}$, SD=$7.86\times{}10^{-2}$; cutoff for outlier weights was set to three SDs). Since the interpretation of the regression weights may depend on possible correlations among the features, we also tested the causal effect of these units on NA-task performance. Ablating the syntax units together resulted in significant performance reduction in NA-tasks that have an interfering noun: Linzen NA-task: $p=0.024$, nounPPAdv-SP: $p=0.011$, nounPPAdv-PS: $p=0.034$, nounPP-SP: $p<0.001$ and marginally significant in nounPP-PS: $p=0.052$ (compared to 1000 random ablations of subsets of units of the same size).

To gain further insight regarding the functioning of the syntax units, we next visualized their gate and cell dynamics during sentence processing. We found that cell activity of unit \unit{2}{500}, which also had one of the highest weights in the regression model, was remarkably structured. The activity of
this unit increases across the entire subject-verb dependency and drops abruptly right after. Figures
\ref{fig:syntax-unit-2Adv} and \ref{fig:syntax-unit-nounpp} show cell activity of this unit during the processing of stimuli from the 2Adv and nounPP tasks. We found the same dynamics in cases where another
verb occurs between subject and main verb, as in subject relatives (Figure \ref{fig:syntax-unit-subjrel}), and in exceptionally long-distance dependencies with two interfering nouns and verbs (Figure \ref{fig:syntax-unit-double-subjrel}). Taken together, these results suggest that unit \unit{2}{500} consistently encodes subject-verb dependencies in a syntax-sensitive manner. Other syntax units did not show an easily interpretable dynamics and had no clear interactions with the number units in the analysis discussed next. This suggests that they perform different syntactic, or possibly other, functions.

\subsection{Syntax-number units connections}

We finally look at the connections that were learned by the LSTM
between syntax unit \unit{2}{500}, which appears to be more closely involved in
tracking subject-verb agreement, and the LR number units, as well as at
the connections between the LR-number units themselves. For each unit pair, there are 4
connection types, one for each component of the target cell (to the 3
gates and to the update candidate). We focus on input and forget gates, as they control the flow and storage of number information.

Figures \ref{fig:interaction-input} and \ref{fig:interaction-forget}
show the distributions of all afferent recurrent weights to the input
and forget gates of the LR-number units, scaled by the maximal activity
$h_t$ of the pre-synaptic units during the nounPP task (this scaling evaluates
the \textit{effective} input to the units and did not change the conclusions described below). We found that the weights from the
syntax unit to the forget gate of both \unit{2}{126} and \unit{2}{338}
are exceptionally high in the positive direction compared to all other
afferent connections in the network ($z-score=8.1, 11.2$, respectively) and
those to their input gates exceptionally negative ($z-score=-16.2,
-7.2$). Since the cell activity of syntax unit \unit{2}{500} is
positive across the entire subject-verb dependency (e.g., Figure
\ref{fig:syntax-unit-double-subjrel}), the connectivity from the
syntax unit drives the number unit forget gates towards one
($W^f_{776, 1150}h^{1150}\gg0$ and $W^f_{988, 1150}h^{1150}\gg0$; $t_{subject}<t<t_{verb}$) and their input gates towards zero
($W^i_{776, 1150}h^{1150}\ll0$ and $W^i_{988, 1150}h^{1150}\ll0$). Looking at the right-hand-side of
Eq.~(\ref{eq:update-rule}), this means that the first term becomes
dominant and the second vanishes, suggesting that, across the entire
dependency, the syntax unit conveys a `remember flag' to the number
units. Similarly, when the activity of the syntax unit becomes
negative at the end of the dependency, it conveys an `update flag'.

Last, we note that the reciprocal connectivity between the two
LR-number units is always positive, to both input and forget
gates (with $|z-score|>3$ for the \unit{2}{126}-to-\unit{2}{338} direction). Since
their activity is negative throughout the subject-verb dependency
(Figures \ref{fig:singular-unit} and \ref{fig:plural-unit}), this means
that they are \textit{mutually inhibiting}, thus steering towards an
unequivocal signal about the grammatical number of the subject to the
output layer.

%
\section{Summary and discussion}
We provided the first  detailed description of the underlying mechanism by which an LSTM language-model performs long-distance number agreement. Strikingly, simply training an LSTM on a language-model objective on raw corpus data brought about single units carrying exceptionally specific linguistic information. Three of these units were found to form a highly interactive local network, which makes up the central part of a `neural' circuit performing long-distance number agreement.

One of these units encodes and stores grammatical number information
when the main subject of a sentence is singular, and it successfully
carries this information across long-range dependencies. Another unit
similarly encodes plurality. These number units show that a highly
local encoding of linguistic features can emerge in LSTMs during
language-model training, as was previously suggested by theoretical
studies of artificial neural networks \cite[e.g.,][]{Bowers:2009} and in
neuroscience \cite[e.g.,][]{Kutter:etal:2018}.

Our analysis also identified units whose activity correlates with syntactic complexity. These units, as a whole, affect performance on the agreement tasks. We further found that one of them encodes the main subject-verb dependency across various syntactic constructions. Moreover, the highest afferent weights to the forget and input gates of both LR-number units were from this unit. A natural interpretation is that this unit propagates syntax-based remember and update flags that control when the number units store and release information.

Finally, number is also redundantly encoded in a more distributed way, but the latter mechanism is unable to carry information across embedded syntactic structures. The computational burden of tracking number information thus gave rise to two types of units in the network,  encoding similar information  with distinct properties and dynamics.

The relationship we uncovered and characterized between syntax and number units suggests that agreement in an LSTM language-model cannot be entirely explained away by superficial heuristics, and the networks have, to some extent, learned to build and exploit structure-based syntactic representations, akin to those conjectured to support human-sentence processing.

In future work, we intend to explore how the encoding pattern we found varies across network architectures and hyperparameters, as well as across languages and domains. We also would like to investigate the timecourse of emergence of the found behaviour over training time.

More generally, we hope that our study will inspire more analyses of the inner dynamics of LSTMs and other sequence-processing networks, complementing the currently popular ``black-box probing'' approach. Besides bringing about a mechanistic understanding of language processing in artificial models, this could inform work on human-sentence processing. Indeed, our study yields particular testable predictions on brain dynamics, given that the computational burden of long-distance agreement remains the same for artificial and biological neural network, despite implementation differences and different data sizes required for language acquisition. We conjecture a similar distinction between SR and LR units to be found in the human brain, as well as an interaction between syntax-processing and feature-carrying units such as the LR units, and plan to test these in future work.

\section*{Acknowledgments}
We would like to thank Kristina Gulordava, Jean-Remi King, Tal Linzen, Gabriella Vigliocco and Christophe Pallier for helpful feedback and comments on the work. 

\bibliography{marco,dieuwke}
\bibliographystyle{acl_natbib}

\end{document}